# Data interference: emojis, homoglyphs, and issues of data fidelity in corpora and their results

**Matteo Di Cristofaro** [1] Università degli Studi di Modena e Reggio Emilia

**Abstract**

Tokenisation – "the process of splitting text into atomic parts" (Brezina & Timperley, 2017: 1) – is a crucial step for corpus linguistics, as it provides the basis for any applicable quantitative method (e.g. collocations) while ensuring the reliability of qualitative approaches. This paper examines how discrepancies in tokenisation affect the representation of language data and the validity of analytical findings: investigating the challenges posed by emojis and homoglyphs, the study highlights the necessity of preprocessing these elements to maintain corpus fidelity to the source data. The research presents methods for ensuring that digital texts are accurately represented in corpora, thereby supporting reliable linguistic analysis and guaranteeing the repeatability of linguistic interpretations. The findings emphasise the necessity of a detailed understanding of both linguistic and technical aspects involved in digital textual data to enhance the accuracy of corpus analysis, and have significant implications for both quantitative and qualitative approaches in corpus-based research.

## 1. Introduction

Corpus linguistics – and quantitative approaches to language data more in general – relies on the ability of computers to count elements in a corpus. Computers may not however always handle this task in the same way a human would, due to variables such as defining what constitutes an element - with corpus linguistics commonly setting a "running word" as the minimal unit of measure, called a 'token' (Brezina, 2018: 39); the language the text is written in; and technical aspects like data format and processing tools. These factors can consequently affect the scientific validity of the results by causing mismatches between source data and the final corpus. 'Tokenisation' - "the process of splitting text into atomic parts" (Brezina & Timperley, 2017: 1) - therefore plays a critical role in ensuring accurate analysis[1], with research showing that different corpus

[1] Recent studies on tokenisation in Large Language Models (LLMs; see also 2.3) have identified similar issues. Although tokenisation in LLMs differs from traditional corpus approaches, both are aimed at splitting text into constituent units; and research has shown that incorrect tokenisation can negatively

tools can produce token quantities that vary by as much as 17% (Brezina & Timperley, 2017). Let us consider Example (1): assuming it is a sentence appearing in the source data used for compiling a corpus, a researcher would count (excluding punctuation) a total of 6 ‘words’ and 5 ‘wordforms’ (or ‘tokens’ and ‘types’ as commonly labelled in corpus linguistics, McEnery & Hardie, 2012: 252). The same researcher would then expect the corpus tool to report the same numbers.

(1) No matter what, love is love!

The same expectation arguably holds true for Example (2), with 8 words and 7 types, as detailed below. The text – a modification of Example (1) - includes two additional features, ‘emojis’ and ‘homoglyphs’, commonly found in so-called ‘computer-mediated communication’ (CMC, i.e. “communication that takes place between human beings via the instrumentality of computers” Herring, 1996: 1).

(2) No matter what, love is 𝕃𝕆𝕍𝔼! 🏳️‍🌈🏳️‍⚧️

Leaving aside interpretations that the use of these two features may lead to (see 2), researchers may argue that for operating corpus methods Example (2) should be analysed as exemplified in Table 1 (columns ***H. tokens*** and ***H. types***). Columns ***C. tokens*** and ***C. types*** report the number of elements as identified by three corpus tools: AntConc[2], LancsBoxX, and SketchEngine (AC, LBX, and SkE henceforth).

| | H. tokens | H. types | C. tokens | C. types |
|---|---|---|---|---|
| No matter what, love is 𝕃𝕆𝕍𝔼! 🏳️‍🌈🏳️‍⚧️ | 8 | 7 | 9 (AC), 13 (LBX, SkE) | 9 (AC), 11 (LBX), 12 (SkE) |

Table 1: Human and computer tokenisation of Example (2)

Mismatches in Table 1 raise both epistemological and practical concerns: first, as the (often sole) ground-truth of a language analysis, a corpus should faithfully represent the language in the source data. ‘Fidelity’ is here intended as absence of discrepancies among a) what the author(s) wrote; b) what the reader see; c) what the source data contains; d) what researchers, once assumed that the previous three are respected, expects the corpus tool to “see”. Mismatches in Table 1 cause researchers to investigate an object (the corpus) not faithful to the language they set out to analyse, thereby undermining the reliability of the theory based on the data (McEnery & Brezina, 2022: 35). What is at stake is not the ability to operate corpus linguistics methods on a corpus, but rather to ensure that these produce results that are faithful to the “linguistic reality” of the source data – i.e. the language under scrutiny. In addition, tokenisation mismatches disrupt core corpus methods like collocations, keywords, n-grams and other descriptive statistics, since they rely on tokens and types (Brezina, 2018). This paper argues that incorrect tokenisation of emojis, homoglyphs, and other UTF-8 features creates ‘sources impact LLM output, hindering precise understanding of input and leading to unsatisfactory results (Wang et al., 2024a).

[2] Using the token definition settings detailed in Table 11

of interference' (SoI henceforth, for both singular and plural form), leading researchers to mistakenly perceive linguistic phenomena caused by unintentional operations (that may therefore go unnoticed) or incorrect readings rather than the actual data contents. As a result, the dataset and procedures underlying the initial interpretation may become unrepeatable – i.e. "reproducing the same results as another researcher given the same data and tools" (McEnery & Brezina, 2022: 25) becomes impossible. To avoid these issues, source data should be investigated and preprocessed prior to corpus compilation; just as "data wrangling" is crucial in statistics (Winter, 2019: 1), it is equally essential in corpus linguistics to manage "digital technicalities" (Di Cristofaro, 2023a:4) influencing how language is created and processed in digital environments. The paper expands on how these digital technicalities affect the creation and analysis of corpus data, and is organised as follows: section 2 reviews how computers process textual data, with a focus on UTF-8, emojis, and homoglyphs in linguistics. Section 3 outlines preprocessing methods and source data. Section 4 presents the analyses, and section 5 provides conclusions and future research directions.

## 2. Humans and computers reading digital textual data

Textual data is read by computers, like any other content in digital format, as sequences of bits – 0s and 1s - subsequently interpreted according to sets of rules called 'encodings' that allow the contents to be rendered. In corpus linguistics, encodings are essential since they ensure that texts are consistently rendered and recognised across different platforms and software, playing a more fundamental role than markup or annotation (Paquot & Gries, 2020: 15). It is outside the scope of this paper to provide an account of the more than 200 existing encodings[3], and this paper only takes into account UTF-8. The next sections describe UTF-8 major underlying technical details (2.1), emojis (2.2), and homoglyphs (2.3), while providing an overview of how these topics have been debated and analysed in linguistics, setting the stage for understanding their role as SoI in corpus data.

### *2.1. Unicode UTF-8 encoding*

The 8-bit Unicode Transformation Format (UTF-8) is the most adopted encoding on the web (98,3% of websites are encoded in UTF-8[4]) - currently (as of July 2024) at version 15.1 released in September 2023. In corpus linguistics it is the suggested encoding for textual data (McEnery & Xiao, 2005) and, together with XML, constitutes "a standard in corpus building" (Paquot & Gries, 2020: 15). It functions like a look-up dictionary, where each string of bits is mapped to a 'codepoint'[5] (representing the bit sequence for a character), which is then mapped to a 'character' (the abstract representation of a symbol). When a text file is opened, the application displays each character as its corresponding 'glyph,' i.e. the visual representation defined by the font. Figure 1 illustrates this process, showing bits, codepoints, and glyphs for two characters; this process applies to all glyphs, including letters, numbers, and other characters supported by the encoding.

[3]For a description and historical overview see Spolsky (2003).

[4]https://w3techs.com/technologies/overview/character_encoding (accessed on 18th July 2024)

[5]A string in the syntax U+N - where N is a sequence of 4 or 5 alphanumerical characters indicating in hexadecimal format the sequence of bits.

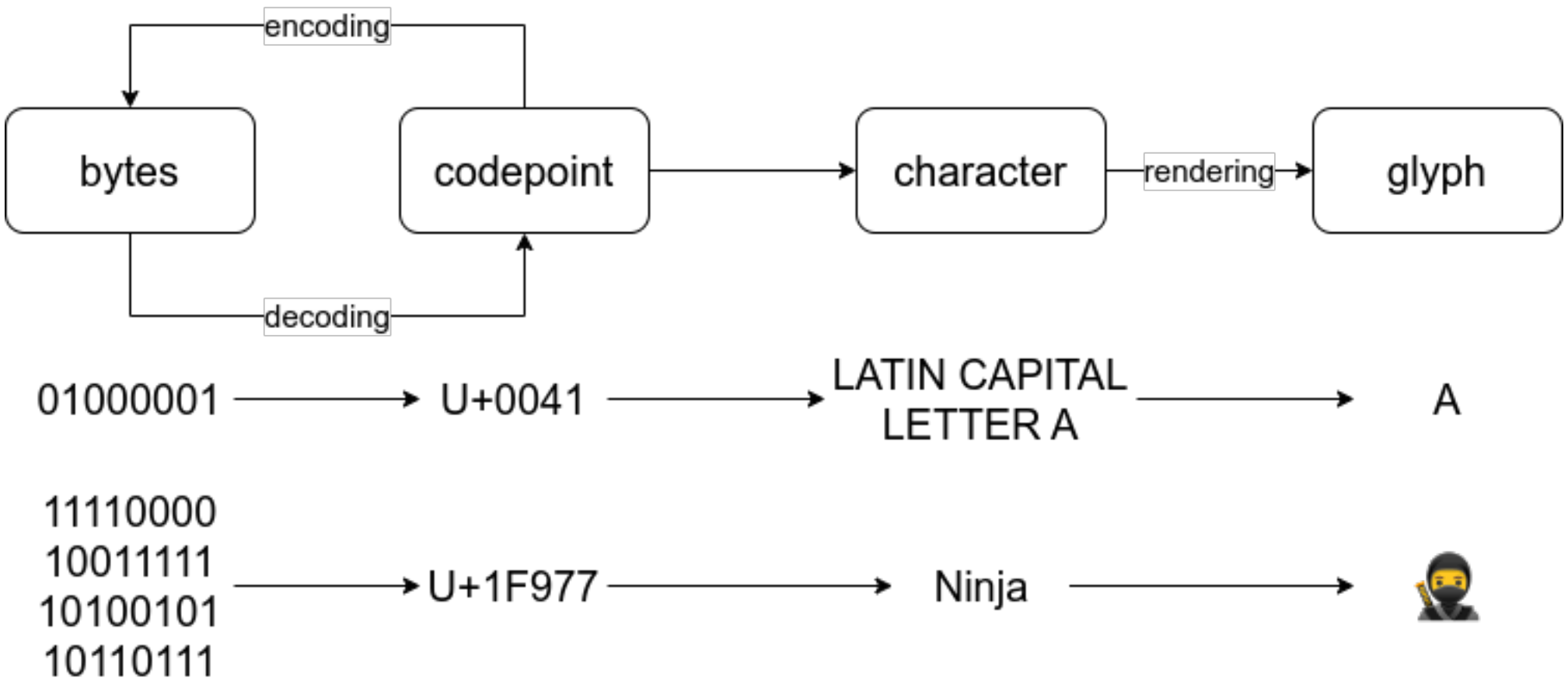

Figure 1: Graph showing computers characters decoding (adapted from Zappavigna & Logi, 2024: 24)

Strings of bits are grouped into sets of eight digits because the “8” in UTF-8 indicates that the minimum number of bits used to encode a codepoint is 1 byte (8 bits), with a maximum of 4 bytes. Table 2 provides examples of characters encoded using 1, 2, 3, and 4 bytes.

| Bits | N. of bytes | Codepoint | Block | Script | Character name | Glyph |
|---|---|---|---|---|---|---|
| 01000001 | 1 | U+0041 | Basic Latin | Latin | LATIN CAPITAL LETTER A | A |
| 11000010 10100011 | 2 | U+00A3 | Latin-1 Supplement | Common | POUND SIGN | £ |
| 11100010 10000010 10101100 | 3 | U+20AC | Currency Symbols | Common | EURO SIGN | € |
| 11110000 10011101 10010100 10111000 | 4 | U+1D538 | Mathematical Alphanumeric Symbols | Common | MATHEMATICAL DOUBLE-STRUCK CAPITAL A | 𝔸 |

Table 2: 1, 2, 3, and 4-bytes UTF-8 characters

Two additional features (‘block’ and ‘script’) are shown in Table 2. A script is defined as “an inventory of graphical symbols, which are drawn upon for the writing systems of particular languages” (Unicode Consortium, 2024): a single script may support many languages, and a language may employ multiple scripts. A block categorises symbols and organises codepoints and characters within the UTF-8 'code space' (the range of hexadecimal addresses assigned to characters). Hence the Basic Latin block contains characters shared across Latin-based writing systems; however, because characters cannot be stored in multiple blocks and each block has limited capacity, block labels can sometimes be misleading (Moran & Cysouw, 2018: 24-25). For example, the currency symbols £ and € are categorized in different blocks, with the former in Latin-1 Supplement and the latter in Currency Symbols (see Table 2). Similarly, the velar nasal *ŋ* used in the International Phonetic Alphabet (IPA) is stored in the Latin Extensions-A block rather than IPA Extensions (Moran & Cysouw, 2018: 25). These peculiarities are linked to challenges in characters processing that stem from the so-called "English bias" (Spence & Viola, 2023), leading to a lack of reliable tools for Non-Latin scripts (NLS). As such NLS complicate the use of NLP tools for both digital versions of historical data or digital-born data, especially when combined with Latin characters (Horváth et al., 2023). Similar issues arise with the romanisation of NLS languages, such as Arabic or Greek (Alsulami, 2019; Lazarinis, 2008), and with multilingual datasets (Eleta & Golbeck, 2014). NLS equally impact web applications, where Latin scripts are often preferred, creating difficulties in processing mixed-script texts (Minnich et al., 2016). Latin-script-only data may nonetheless be affected by such issues, particularly when non-letter characters – numbers, symbols, emojis - are used to e.g. avoid censorship (Calhoun & Fawcett, 2023). Consequently, the issues discussed in this paper potentially apply not only to emojis and homoglyphs but also to ligatures, accented characters, and NLS, all of which may interfere with tokenisation. Adding further complexity to textual data processing are operating systems-related aspects, such as non-printable characters (Python Software Foundation, 2024, str.isprintable() section): for instance, Windows uses CR + LF for line breaks, while Unix-based systems use LF alone, complicating file exchanges between systems[6]. Windows has also historically adopted UTF-16 based encodings (Chen, 2019) - with UTF-8 support introduced in Windows 10 (Bridge et al., 2023) -, whereas Unix-based ones default to UTF-8 (Gries, 2016: 111). Due to these complexities, the data and scripts used in this paper are processed on a Unix-based system with UTF-8 encoding and LF for line breaks[7] to minimise potential technical issues.

### 2.2. *Emojis*

As a result of the widespread use of emojis in digital communications (Logi & Zappavigna, 2021) a substantial body of research has focussed on this feature over the past 10 years. Research on emojis - originally derived from studies on emoticons (Zappavigna & Logi, 2021: 2) - has evolved rapidly, documenting their flexible usage as carriers of varied (Fricke et al., 2024) and indirect meanings (Holtgraves & Robinson, 2020). Studies have also observed how a number of emojis tend to acquire conventionalised meanings (Weissman et al., 2023), with others "becoming pragmatically ***unmarked***" (Konrad et al., 2021: 229; emphasis in the original) or semantically ambiguous (Hafner, 2023), undergoing processes of grammaticalisation (Wiese & Labrenz, 2021). Despite these advances, studies using quantitative approaches have often overlooked emojis technical challenges, or have left mismatches between source data and corpora unaddressed (Zappavigna & Logi, 2021; Wang et al., 2024b). Exceptions include work by Shoebu and de Melo (2021) and Zappavigna and Logi (2024); with a more general overview together with practical code on how to process emojis in Di Cristofaro (2023a: 283-285, 344-346) - which this paper extends and complements. Most linguistic studies have however agreed upon "what" and emoji is, with the aforementioned studies – along with the majority of existing ones – considering it as a single pictorial image accessed via emoji keyboards or palettes (Davis & Holbrook, 2023). Based on this understanding the string in Example (3) contains three emojis.

[6] Advanced text editor (e.g. Geany) allow to manually set the line break independently of the operating system in use.

[7] OpenSUSE Tumbleweed with en_GB.UTF-8 locale; readers employing Windows machines to browse and run the accompanying materials need to verify they are using UTF-8 – or should consider using a (virtual) Unix machine for working with textual data (Gries, 2016: 111).

(3) 😵‍💫👩‍❤️‍💋‍👩🤹🏽‍♀️

This way of counting underpins a 'human-centred' perspective, always coexisting with a 'computer-centred' one: a number of emojis that a researcher would count as single characters are in fact processed by computers as multiple characters, giving rise to the mismatches reported in Table 1. Table 3 shows for each emoji in Example (3) – columns ***Emoji***, first row – its codepoints, with subsequent rows showing both graphical depiction and codepoint for each emoji character that compose the ones in the first row. Empty graphical representations are explained further below.

| Emoji | Codepoint | Emoji | Codepoint | Emoji | Codepoint |
|---|---|---|---|---|---|
| 😵‍💫 | U+1F635<br>U+200D<br>U+1F4AB | 👩‍❤️‍💋‍👩 | U+1F469<br>U+200D U+2764<br>U+FE0F U+200D<br>U+1F48B<br>U+200D<br>U+1F469 | 🤹🏽‍♀️ | U+1F939<br>U+1F3FD<br>U+200D U+2640<br>U+FE0F |
| 😵 | U+1F635 | 👩 | U+1F469 | 🤹 | U+1F939 |
| | U+200D | | U+200D | 🏽 | U+1F3FD |
| 💫 | U+1F4AB | ❤ | U+2764 | | U+200D |
| | | | U+FE0F | ♀ | U+2640 |
| | | | U+200D | | U+FE0F |
| | | 💋 | U+1F48B | | |
| | | | U+200D | | |
| | | 👩 | U+1F469 | | |

Table 3: Analysis of emojis in Example (3)

To develop a procedure that minimises SoI, it is essential to first establish labels for discussing the relevant technical details. According to Unicode, an emoji is defined as a "colorful pictograph that can be used inline in text [...] [and is internally repre-sented as] either (a) an image, (b) an encoded character, or (c) a sequence of encoded characters" (Davis & Holbrook, 2023, Definitions section). To encompass these three options, while employing the general term 'emoji' to indicate the aforementioned human-centred perspective, the following terms - adapted from Davis and Holbrook (2023) – are adopted:

1. 'glyph' refers to (a); in this paper it is a synonym of emoji, as the three emojis reported in Example (3). Different 'emoji images' may however appear to different readers (see further below).
2. 'emoji character' refers to (b); e.g. the two emojis 😵 ("Face with Crossed-Out Eyes") and 🤹 ("Person Juggling") are internally represented by a single emoji character each (see Table 3)
3. 'emoji sequence' refers to (c); e.g. the emoji 😵‍💫 ("Face with Spiral Eyes") is internally composed of the two emoji characters 😵 and 💫 ("Dizzy") joined by the Zero-Width Joiner component (see further below)
4. 'emoji modifier' refers to those "character[s] that can be used to modify the appearance of a preceding emoji" (Davis & Holbrook, 2023, Emoji modifiers section); e.g. in the emoji sequence 🤹🏽‍♀️ ("Woman Juggling: Medium Skin Tone") the Medium Skin Tone is assigned through the Fitzpatrick modifier[8] 🏽 ("Medium Skin Tone")
5. 'emoji component' refers to those characters appearing in sequences but not appearing in emoji keyboards as separate choices. Examples of components are the Zero-Width Joiner (ZWJ; codepoint: U+200D), 'Variation Selectors', and 'Regional Indicators'. The latter, if split as separate characters, are "displayed as […] capital A..Z character[s] with a special display" (Davis & Holbrook, 2023, Conformance section) such as 🇬 (codepoint: U+1F1EC) and 🇦 (codepoint: U+1F1E6). These are used to produce i.a. glyphs of regional flags; in this case the Gabon flag (🇬🇦; codepoint: U+1F1EC U+1F1E6). Variation Selectors include different characters, such as Variation Selector-15 (codepoint: U+FE0E) and Variation Selector-16 (codepoint: U+FE0F) used to indicate that the emoji character should be displayed as either text or as an emoji glyph[9] respectively. ZWJ is used "between the elements of a sequence of characters to indicate that a single glyph should be presented if available. […] [If not] a fallback sequence of separate emoji is displayed" (Davis & Holbrook, 2023, Emoji ZWJ Sequences section).

Emoji images can differ visually depending on the device used (Zappavigna & Logi, 2024: 28-30), as various vendors (such as Google or Microsoft) design their own sets of emoji images in alignment with Unicode recommendations. Vendors may however deviate from the guidelines, leading to significant differences in how emojis are rendered: for instance, the depiction of professions such as police officer or detective varies in terms of gender representation across platforms (Davis & Holbrook, 2023)[10]. Research on how graphical differences in emojis affect digital communication is limited (Zappavigna & Logi, 2024: 30), but suggests that they lead to ambiguity and miscommunication (Shurick & Daniel, 2020; Miller et al., 2021), often due to users' unawareness of the different visual representations (Miller Hillberg et al., 2018). This presents challenges for researchers, as it may be difficult to determine how users interpret or employ emojis without involving them directly in studies - as in Shurick and Daniel (2020). An issue that might however be mitigated if metadata documenting the platform or application used by the user is available in the source data.

[8]Based on the Fitzpatrick scale, "a recognized standard for dermatology" (Davis & Holbrook, 2023, Diversity section), five types of emoji modifiers exist: Light, Medium-Light, Medium, Medium-Dark, and Dark skin tone.

[9]Examples of Variation Selector usage can be found at https://www.ii.com/unicode-variation-selectors-15-16/ and are not included in this paper since document editors (Word, LibreOffice Write) do not show the variants as different.

### 2.3. *Homoglyphs*

Homoglyphs are "visually indistinguishable glyphs (or highly similar glyphs) that have different codepoints in the Unicode Standard and thus different character semantics" (Moran and Cysouw, 2018: 26), such as the ***A*** characters in Table 2, or the ones used to write the occurrences of love in Example (2). While their linguistic significance has often been overlooked, homoglyphs have been extensively studied in cybersecurity[11], and in Natural Language Processing (NLP) since they can disrupt tools for hate speech detection and machine translation (Cooper et al., 2023). Recent advancements in Large Language Models (LLMs)-based AIs have also drawn attention to homoglyphs' potential to cause biases in AI-generated content, including image generation errors (Struppek et al., 2023) and comprehension mistakes (Daniel & Pal, 2024). Despite the centrality played by language (both as input and as output) in LLMs (Gillings et al., 2024: 5), and the prominence that LLMs are acquiring in various fields of linguistics - from corpus linguistics (i.a. Curry et al., 2024; Uchida, 2024), to language teaching and learning (Kohnke et al., 2023), Discourse Studies (Gillings et al., 2024), and move analysis (Yu et al., 2024) -, little focus has been devoted to linguistic aspects of homoglyphs. A particularly concerning situation considering the widespread usage of homoglyphs in social media platforms, and the role that 'typefaces' (i.e. the visual design of letters, numbers, and symbols) play in the construction of (digital) brands and identities (i.a. McCarthy & Mothersbaugh, 2002; Andrade et al., 2024). Addressing these issues may therefore require a shift toward incorporating the technical aspects of homoglyphs into linguistic studies and LLM training; in order to (hopefully) facilitate these efforts, and to better understand the role that homoglyphs play as SoI in corpus data, some background details on their technical aspects are need. Unicode defines two types of equivalence between characters: 'canonical' and 'compatibility'. Canonical refers to characters or sequences of characters that "when correctly displayed should always have the same visual appearance and behavior", while compatibility accounts for characters that share the same abstract meaning but differ visually or behaviourally (Whistler, 2023). Table 4 exemplifies four pairs of character(s) equivalences, indicating for each the type and subtype of equivalence (adapted from Whistler, 2023, where the full list may be found).

[10]Examples of variations may be found at https://emojipedia.org/vendors

[11]Homoglyphs are used in various attacks, such as impersonating characters in URLs to create fake websites (Suzuki et al., 2019), or to hide encrypted messages in social media (Bertini et al., 2019)

| Character(s) | Character(s) | Equivalence type | Equivalence subtype |
|---|---|---|---|
| é | e+◌́ | Canonical | Combining sequence |
| q+◌̇+◌̣ | q+◌̣+◌̇ | Canonical | Ordering of combining marks |
| ℌ | H | Compatibility | Font variants |
| ① | 1 | Compatibility | Circled variants |

Table 4: Examples of character equivalence types

Unicode defines procedures called Unicode Normalization Forms (UNF) that allow applications to process characters by combining or decomposing sequences, with four main forms, summarised in Table 5. These vary in how they handle character 'composition' and 'decomposition' (Whistler, 2023): the former combines sequences of codepoints into precomposed characters (e.g. *é*), the latter decomposes precomposed characters (e.g. e+◌́). Fig.2 demonstrates the results of each UNF applied to three different characters.

| Form | Description |
|---|---|
| Normalisation Form D (NFD) | Canonical Decomposition |
| Normalisation Form C (NFC) | Canonical Decomposition, followed by Canonical Composition |
| Normalisation Form KD (NFKD) | Compatibility Decomposition |
| Normalisation Form KC (NFKC) | Compatibility Decomposition, followed by Canonical Composition |

Table 5: Unicode Normalization Forms (Whistler, 2023)

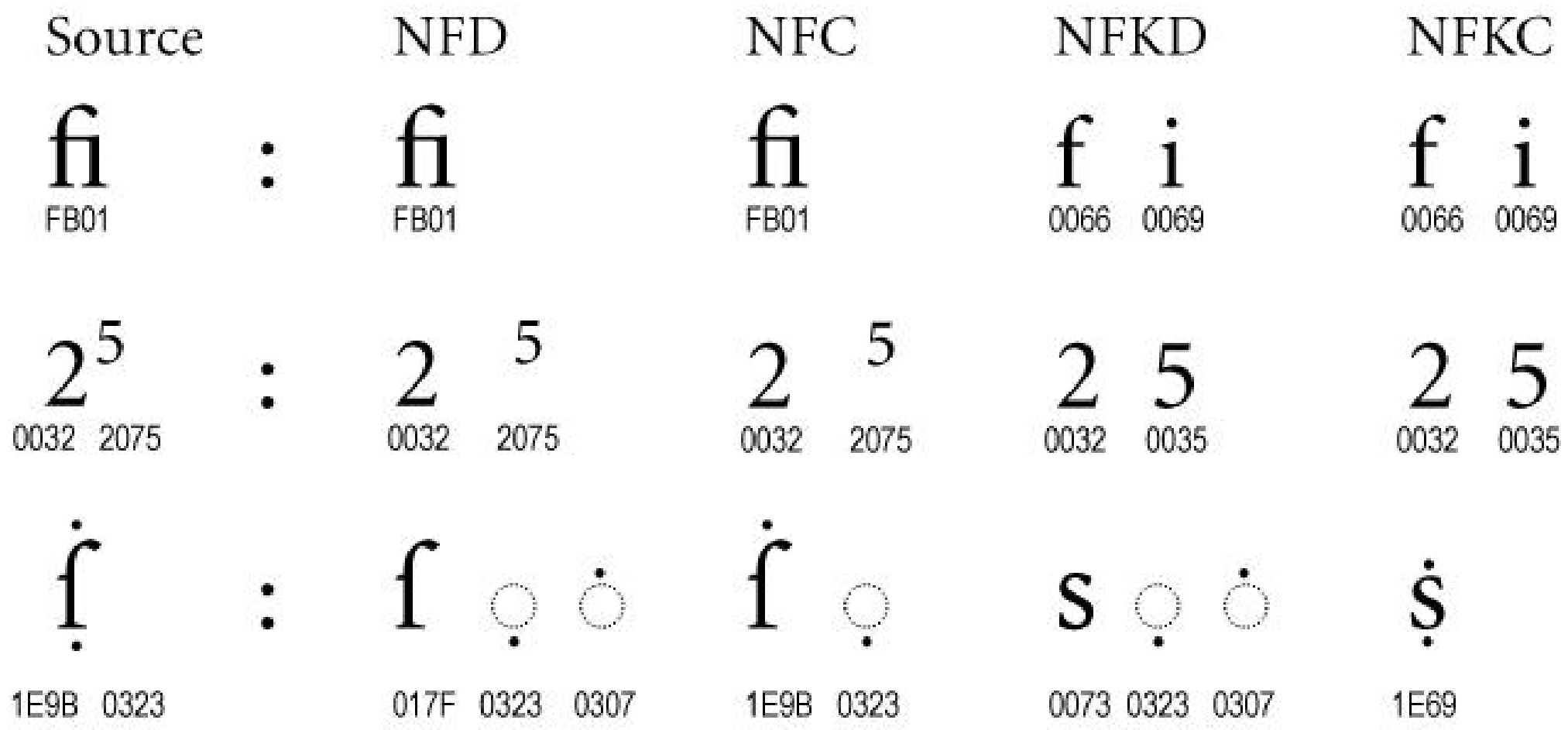

Figure 2: Examples of compatibility composites (Whistler, 2023)

Each UNF may serve different purposes and, depending on how the data needs to be processed, "a software application may compose, decompose or reorder characters as its developers desire [...] as long as the resultant strings are to the original" (Moran & Cysouw, 2018: 29). For example the TEI guidelines suggest NFC (TEI Consortium, 2023: lii), whereas Moran and Cysouw (2018: 35) suggest either NFC or NFD. It is generally accepted that normalisation forms that operate compatibility decomposition may irreversibly alter the source data (e.g. the digit ***5*** in Fig.2 has a different meaning whether it is subscripted or not), thus making both NFKC and NFKD forms to be considered "destructive" (Phillips 2021). However it is also true that the results produced by these forms may facilitate the recognition of equivalent strings of characters for the purpose of corpus linguistics in cases where the same abstract character has different visual styles – additionally ensuring that each "abstract character [is] represented in one way only in a given Unicode document or document collection" (The TEI Consortium, 2023: lii). Hence, and despite current suggestions, the present paper adopts NFKC, based on three considerations: i) NFKC and NFKD both guarantee that multi-glyphs homoglyphs are always normalised into their single normalised components (see ligature in Fig.2); ii) NFKC always normalises multi-component homoglyphs to single precomposed characters, whereas NFKD does not (see character *ş* in Fig.2); iii) source data information regarding homoglyphs destroyed during the normalisation procedure can be preserved through metadata, e.g through XML as in Example (4).

(4) `No matter what, love is <norm orig="𝕃𝕆𝕍𝔼">love</norm>!`

Details discussed so far, along with the presented suggestions, are aimed at ensuring that, independently of the corpus tool used, two equivalent abstract characters are correctly recognised as such. There are however a number of abstract characters whose glyphs, depending on the rendering font, may look similar (or identical) to the human eye; an example is the similarity of the digit ***1*** with the LATIN SMALL LETTER L ***l*** when using Times New Roman. To prevent such cases Unicode defines a list of 'confusable'[12] glyphs in its Unicode Security Mechanisms (Unicode Consortium, 2023); these however are not taken into account since they do not represent a potential SoI, but rather are dependant on the selection of the rendering font, thus impinging on the human eye only.

## 3. Methodology

Corpus tools often use custom rules for processing text files, which can affect the handling of UTF-8 features like emojis and homoglyphs. This paper stems from the observation of how three corpus tools (AntConc, LancsBoxX, and SketchEngine) operate differently on the same initial dataset (see 3.2), leading one set of non preprocessed corpus files to generate mismatches in the resulting corpus. While variations in tokenisation across tools are attested (see Introduction), the nature of the observations investigated here is less documented. It must here be stressed that observations do not arise from a comparison among the different tools, but rather from an intra-evaluation between the source data and the resulting corpus. The analyses aim to: i) evaluate the impact of emojis and homoglyphs preprocessing on the final corpus and highlight the effects of non-preprocessed data on results, and ii) describe how different tools handle non-preprocessed emojis and homoglyphs, offering guidance on configuring tools and preparing data for accurate analysis. The tools used - AntConc v4.2.4 (Anthony, 2023), LancsBoxX v4.0.0

[12] https://www.unicode.org/Public/security/15.1.0/

(Brezina & Platt, 2024), and SketchEngine (Kilgarriff et al., 2014[13] - were chosen for their popularity and ease of use, not requiring server setups[14]. Initial testing involves "artificial data" consisting of isolated emojis and Latin/Common script characters (3.1). "Real data" (3.2) is then prepared (3.3) to identify further tools' behaviours when emojis and homoglyphs occur in natural language. All the analyses and considerations, being based on English data tagged using English taggers (this is valid for both LBX and SkE), are limited to a subset of the existing Unicode characters, despite the presence of Non-Latin scripts in the source data[15]. All materials – source data, corpus files, scripts and Jupyter notebooks used for data preprocessing and evaluations - are available at [note: the link will be made available in the published version].

### 3.1. *Emojis and homoglyphs test files*

Identification of emojis for the three corpus tools was verified by creating a test file containing all 5,034 emojis supported by the Python library `emoji` (version 2.12.1; Kim, 2024), subsequently compiled into a corpus in each corpus tool. The list[16] - containing all 3,782 Unicode v15.1 emojis plus more than 1,000 aliases (Kim, 2024) - was converted into a tab-separated value file: each emoji is written on a single line, preceded by the string ***emoji*** separated by a tab character and followed by a newline – as in Example (5), where the → symbol indicates a tab character. Evaluation is conducted through wordlists comparisons, as if correctly tokenised each emoji would appear as a type with a frequency of 1.

(5)
emoji→👩🏻
emoji→💙

A similar procedure was adopted for homoglyphs, where one row for each character is created – the character is preceded by the string ***character*** followed by a tab, and is followed by a newline: 8,160 characters (types and tokens, since every character is unique) are included, i.e. all the UTF-8 characters that i) are considered as printable by Python[17] and ii) are included in either the Latin or Common script. Since SkE operates NFKC homoglyphs normalisation, an NFKC normalised test file[18] was created to evaluate its results; this contains 8,644 characters (tokens) and 5,850 types, since a number of homoglyphs once normalised produce the same glyph (e.g. 𝕃, ℒ, and ℓ are all normalised to ***l*** ), and others produce multiple ones (e.g. ⒜ produces ***(***, ***a***, and ***)*** ).

---

[13] SketchEngine does not provide version numbers; all corpora and results were compiled and obtained between 1st and 25th July 2024.

[14] WordSmith Tools 9 was initially considered but discarded due to its reliance on UTF-16 LE rather than UTF-8 (see https://lexically.net/downloads/version9/HTML/flavours_of_unicode.html), which caused emojis to be unprocessed and resulted in empty wordlists, making comparisons ineffective. Although conversion to UTF-8 is possible, it was excluded to avoid adding complexity, as all source data and most web data are encoded in UTF-8 (see 2.1).

[15] The following scripts were found, with the number of characters and the total frequency in parentheses: Braille (1; freq.: 6), Han (16; freq.: 24), Hiragana (23; freq.: 44), Katakana (21; freq.: 30).

[16] The list is contained in the dictionary employed by the Python library `emoji` (https://github.com/carpedm20/emoji/blob/e83cf4acf6de5e5918b9a34f079214cde2aab87c/emoji/unicode_codes/data_dict.py), and is compiled from the Unicode Standard.

[17] According to Python 3.12 built-in method `isprintable()`, https://docs.python.org/3/library/stdtypes.html#str.isprintable .

[18] Normalisation was conducted through the Python embedded module unicodedata, which uses official Unicode Consortium data; see https://docs.python.org/3/library/unicodedata.html .

### 3.2. *Source data*

Source data comprises all the posts (without comments) published on Instagram by 13 fashion brands between 1st January and 31st December 2022 (see Table 12 for further details). This is a subset of a larger dataset for the project ***Communicating transparency - New trends in English-language corporate and institutional disclosure practices in intercultural settings (PRIN 2020TJTA55)***, containing Instagram data collected across the same time period for companies in the fashion and the transportation market sectors. Thirteen brands were selected from the ***Transparency-Fashion*** corpus, containing 102 fashion brands and 3.3 million tokens (Di Cristofaro, 2023b): files for one random brand were selected at a time, their token counts recorded, and an additional brand was selected until the total exceeded 200,000 tokens[19]. This limit was set arbitrarily to ensure the corpus could be manually investigated and analysed. Data was collected through Instagram scraper `instaloader`[20] using the command in Example (6), which collects all posts between the two set dates from the defined `TARGET_ACCOUNT` using `USER_ACCOUNT` login details, while excluding the retrieval of multimedia files (`--no-pictures` and `--no-videos`). Further details as to how `instaloader` data is saved and structured may be found in Di Cristofaro (2023a: 208).

(6)
```
instaloader --login=USER_ACCOUNT --no-pictures --no-videos –post-filter="date_utc >= datetime(2022,1,1) and date_utc <= datetime(2022,12,31)" TARGET_ACCOUNT
```

### 3.3. *Data preprocessing procedures*

Preprocessing of both emojis and homoglyphs to mitigate the effect of SoI is achieved through two different procedures; these are the result of repeated testing and refinement on the source data, its characteristics, and the linguistic purposes of the corpus. Emojis preprocessing includes two different steps: first (‘retokenisation’) a whitespace is added where one or more emojis is present in the text and not separated with a whitespace from preceding or following words or emojis (similar to the procedure adopted by Shoeb and de Melo, 2021), as in Example (7) - where the first row shows the original text, and the second the retokenised one; character ‘|’ represents a whitespace. The use of the library `emoji` guarantees that emoji characters and sequences are correctly interpreted and separated from the surrounding elements.

(7) 🛼🛼down
🛼 | 🛼 | down

Text with added whitespaces is then fed to PyMUSAS (Moore & Rayson, 2022), where the word(s) and each preceding or following emoji are tokenised as separate elements. The second step (‘transliteration’) transforms emojis into their official textual description

[19]Tokens are calculated using PyMUSAS on the source data processed through the procedure described in 3.3.

[20]https://instaloader.github.io/

(‘CLDR short names’; Davis & Holbrook, 2023, Names section) using the library `emoji`. This procedure expands on the one outlined in Di Cristofaro (2023a: 283-285), and is exemplified in Example (8) - where the first row shows the result from retokenisation, and the second row the transliterated version; character ‘|’ indicates token division.

(8) 🛼 | 🛼 | down
{roller^skate} | {roller^skate} | down

Non-alphanumeric characters in the CLDR name are replaced with a caret (^) and the resulting short name is enclosed in curly brackets {}[21]. These replacements prevent the tool from misinterpreting transliterated emojis and splitting them into different words. The characters *^*, *{*, and *}* were chosen after confirming that they did not appear in the source data; during corpus compilation they were added to “Token Definition” (as named in AntConc) list of characters so that e.g. ***{roller^skate}*** is tokenised as one single token. Emoji transliteration was performed prior to NFKC normalisation to prevent the transformation of 31 emojis (22 unique and 9 variants) that could be normalised through NFKC or NFKD[22]. This decision is arbitrary, and based on the desiderata of querying e.g. the emoji ™ (codepoint: U+2122) as the element ***{trade^mark}*** rather than the string ***TM*** – the latter would result were the NFKC normalisation be applied before transliteration. For homoglyphs normalisation, every token identified by PyMUSAS is run through a function that checks whether the textual content of the token is already normalised. If not, NFKC is applied and the result is tokenised once again through PyMUSAS – since, once normalised, it may result in two or more different tokens. At last each final token is added to the contents of the preprocessed corpora files. Example (9) reports a sample input (top row) and output (bottom row) of the normalisation – were the ‘|’ character indicates token division.

(9) **PRESIDENT’S**
PRESIDENT | ‘S

Following the results of preliminary analyses, corpus files were saved into three different formats (.txt, .xml, and .vrt) to accommodate each corpus tools features: whereas .txt is compatible with all three tools, neither LBX nor SkE allow for the definition of custom token characters, and are therefore not able to treat the transliterated emojis enclosed in curly brackets as single tokens. Contents of the preprocessed files were therefore tagged (POS and lemma) using PyMUSAS[23] and subsequently saved in .xml (for LBX) and .vrt (for SkE). Table 6 summarises the different corpus files sets created.

[21] Searching for all emojis in the preprocessed corpus requires the following syntaxes: `\{.*?\}` (AC, using Regex query); `[word="\{.*?\}"]` (LBX); `[word="\{.*\}"]` (SkE, using CQL query).

[22] The list is available in the accompanying materials.

[23] PyMUSAS v0.3.0 with language modules `en-core-web-sm v3.7.1` and `en-dual-none-contextual v0.3.3`

| AntConc | | LancsBoxX | | SketchEngine | |
|---|---|---|---|---|---|
| *procedure* | *format* | *procedure* | *format* | *procedure* | *format* |
| Non pre-processed | txt | Non pre-processed | txt | Non pre-processed | txt |
| Preprocessed | txt | Preprocessed | xml | Preprocessed | vrt |

Table 6: Corpus files sets created for the three tools

## 4. Consequences of SoI in corpora and their results

It is important to remark once again that the aim of the paper is not to suggest "which corpus tool is best", but rather to discuss a number of their functionalities in relation to the treatment of SoI in corpus files, and elucidate what should be considered when compiling or analysing a corpus, while providing evidence of what happens when no preprocessing is operated. The analyses are consequently structured as follows: first each tool's behaviour is evaluated using "artificial data" (3.1) for emojis (4.1) and homoglyphs (4.2). Then the prepared corpus files (3.3) are employed to identify further tools' behaviours dependant on the use of emojis and homoglyphs in occurrences of natural language (4.3).

### *4.1. Emojis detection in corpus tools*

| Corpus tool | N. emoji types | N. emoji types freq. = 1 | N. emoji types freq. > 1 |
|---|---|---|---|
| AntConc[24] | 2,489 | 2,088 | 401 |
| LancsBoxX | 1,627 | 1,191 | 436 |
| SketchEngine | 3,872 | 3,608 | 264 |

Table 7: Emojis detection results

Results from the emojis procedure described in 3.1 are reported in Table 7: number of emoji types identified by each tool, number of the ones with a frequency of 1, and of those with a frequency > 1. Manual analysis of the identified types shows that not all of them represent valid emojis, and as such it can be stated that none of the corpus tools under consideration are able to correctly tokenise all current (v15.1) emojis, albeit with substantial differences. AC is not able to process emojis sequences where the emoji component ZWJ (see 2.2) is used, leading ZWJ sequences to be split into their separate elements (emoji characters or sequences without ZWJ), and ZWJ to be excluded from tokenisation. LBX similarly does not correctly tokenise ZWJ sequences, leading however to mixed results: ZWJ is itself tokenised (freq. = 931), and also appears in other 87 types – the majority of which are invalid emoji sequences, resulting from splitting valid sequences into invalid ones. SkE also presents a similar behaviour, with ZWJ appearing as a type (freq. = 1,639) and in other 1,502 types including invalid sequences. Contrary to LBX however, SkE does not split Fitzpatrick Modifiers (see 2.2) from their relative person emojis (this is only true for emojis in Unicode version < 13), while the former does. Looking at the results from the perspective of Unicode versions, more details emerge: Table 8 reports how many emoji types are tokenised by each tool according to each Unicode version – with the total number of existing emojis for each version in row ***Unicode***. Values emphasised in bold indicate that the tool identifies the exact number of emojis included in that specific version of Unicode.

[24]Emoji test file was loaded by adding to the default settings the categories "Symbols" (sub-categories "Modifier" and "Other") and "Marks". The "Symbols" is required to count emojis as tokens (as per Laurence Anthony's communication, https://x.com/antlabjp/status/1498980318849941512 ),while the "Marks" is required to count the variant selector U+FE0F.

| | v0.6 | v0.7 | v1 | v2 | v3 | v4 | v5 | v11 | v12 | v12.1 | v13 | v13.1 | v14 | v15 | v15.1 | Inv. |
|---|---|---|---|---|---|---|---|---|---|---|---|---|---|---|---|---|
| Unicode | 793 | 254 | 512 | 297 | 157 | 1 030 | 339 | 188 | 266 | 186 | 146 | 422 | 112 | 31 | 301 | — |
| AntConc | 749 | **254** | **512** | 264 | **157** | 22 | 136 | 90 | 101 | 0 | 67 | 10 | 92 | 30 | 0 | 5 |
| LancsBoxX | 723 | 139 | 175 | 1 | 72 | 3 | 56 | 68 | 86 | 0 | 66 | 0 | 107 | 30 | 0 | 101 |
| SketchEngine | 740 | **254** | **512** | **297** | **157** | **1,030** | 336 | **188** | **266** | 0 | 56 | 0 | 0 | 0 | 0 | 336 |

Table 8: Emojis detection results by Unicode version

Results from Table 8 confirm the incorrect tokenisation of ZWJ sequences by AC and LBX, made evident by the number of emojis recognised for v15: the only missing is the one where ZWJ is present – 🐦‍⬛ ('Black Bird'; codepoint: U+1F426 U+200D U+2B1B). Column ***Inv.*** reports the number of tokenised characters or sequences that do not constitute valid emojis but rather emoji components such as ZWJ or Regional Indicators (see 3.1), or invalid sequences. The presence of invalid emojis also indicates that tools revert to fallback sequences of separate emojis (see 2.2) and then splits them, and that none of the tools currently (as of July 2024) support v15.1 emojis such as "direction-specifying versions"[25] - resulting in their component characters to be counted as separate tokens.

### 4.2. *Homoglyphs detection in corpus tools*

| Corpus tool | Types | Types freq. > 1 | Tokens | Missing types | Missing tokens |
|---|---|---|---|---|---|
| SketchEngine | 5,755 | 637 | 8,252 | 95 | 392 |
| LancsBoxX | 7,272 | 508 | 7,786 | 888 | 374 |
| AntConc | 2,000 | 466 | 2,476 | 6,160 | 5,684 |

Table 9: Homoglyphs detection results

Results for the homoglyphs procedure described in 3.1 are presented in Table 9, showing the number of types (unique homoglyphs) and tokens (occurrences of all the types) identified by each tool under scrutiny. Columns ***Missing types*** and ***Missing tokens*** report the difference for types and tokens as counted by each tool, against the total number of types and tokens in the original test file. Difference for SkE was calculated from the normalised test file, whereas the non-normalised test file was used for comparison with the remaining tools (see 3.1). Despite SkE operating normalisation, this appears to be selective or partial: only two examples are presented to illustrate the potential cause, since it is outside the scope of this paper to list all the differences this produces.

[25] https://blog.emojipedia.org/whats-new-in-unicode-15-1-and-emoji-15-1/

Mismatches between SkE and the test file for the types ***c*** (SkE freq.: 44; test file freq.: 45) and ***q*** (SkE freq.: 35; test file freq.: 37) indicate that the former does not apply any normalisation to characters from blocks Latin Extended-D and Latin Extended-F; Table 10 contains the ***c*** and ***q*** homoglyphs that SkE does not normalise.

| Non-normalised homoglyphs | Codepoint | Block |
|---|---|---|
| ꟴ | U+A7F4 | Latin Extended-D |
| 𐞥 | U+107A5 | Latin Extended-F |
| ꟲ | U+A7F2 | Latin Extended-D |

Table 10: Homoglyphs non-normalised by SketchEngine

Expanding the analysis to all Latin Extended blocks (from A to G), it appears that the non-normalisation (or complete exclusion from tokenisation, as in the case of characters from Latin Extended-G[26]) of these characters is deliberate, since out of the total 711 characters across all Latin Extended blocks, only a minority (n = 86) of them can be normalised using NFKC. As for AC and LBX, neither one applies any sort of normalisation, but the former gives the possibility of defining tokenisation options. Consequently different results are produced by AC depending on the ***Token Definition*** options applied: 2,473 types and 10,632 tokens are identified using the ***Default*** settings; 7,547 types and 15,706 tokens when using all character-relevant ones[27]. When using AC with data that contains homoglyphs, options should consequently be selected on a per-case basis, starting from an evaluation of which homoglyphs are included in the data. A number of mismatches are however produced by both AC and LBX: while upper and lower case versions of the same character are recognised, all ligatures such as ***ﬁ*** (codepoint: U+FB01) are counted as types, and are thus not searchable by using the query ***fi*** (codepoints: U+0066 U+0069); or similarly all 18 homoglyphs for each LATIN CAPITAL LETTER appear as 18 different types.

### *4.3. SoI tokenisation in natural language data*

Corpus files sets (see 3.3) were at this stage compiled into corpora in the three different tools for the calculation of wordlists. Both in LBX and SkE non preprocessed and preprocessed files were loaded using default settings; in AC different ***Token Definition*** settings reported in Table 11 were adopted.

[26]These are mostly meant for phonetic transcriptions, see https://unicode.org/charts/nameslist/n_1DF00.html

[27]i.e. option groups labelled as 'Unicode Punctuation', 'Custom Punctuation', 'Marks', and 'Symbols'

| non preprocessed files | | preprocessed files | |
|---|---|---|---|
| Token category | Defined tokens | Token category | Defined tokens |
| Letters | Uppercase; Lowercase; Modifier; Other | Letters | Uppercase; Lowercase; Modifier; Other |
| Numbers | Decimal | Numbers | Decimal[28] |
| Symbols | Other | User-Appended Characters | ^{} |

Table 11: AntConc 'Token Definition' settings

Table 12 reports details for each corpus file (one file for each brand, see 3.2): along with the name of the ***brand*** and the Instagram account (***IG handle***) from which the data was collected, also included are the number of Instagram posts that compose each corpus file; the number of types and tokens as identified by PyMUSAS on the preprocessed version (including emojis and normalised homoglyphs); the total number of emojis as recognised by the library `emoji`; and the total number of homoglyphs (as recognised by the library `unicodedata`) as well as their resulting tokens once normalised using NFKC (***NFKCNorm***).

[28]Decimals are required to avoid splitting transliterated emojis that have digits in their short name, such as ***keycap^1***

| Brand | IG handle | N. posts | Types | Tokens | Emojis | Homoglyphs | NFKC-Norm |
|---|---|---|---|---|---|---|---|
| Calzedonia | @calzedonia | 563 | 1,223 | 8,985 | 476 | 2 | 2 |
| Carolina Herrera | @CarolinaHerrera | 1,074 | 4,378 | 51,128 | 985 | 56 | 59 |
| Champion | @champion | 95 | 734 | 1,654 | 51 | 3 | 3 |
| Chloé | @chloe | 500 | 3,062 | 20,049 | 7 | 11 | 32 |
| Columbia Sportswear | @columbia1938 | 177 | 1,626 | 5,723 | 141 | 10 | 10 |
| DKNY | @dkny | 345 | 1,101 | 5,023 | 53 | 0 | 0 |
| Forever21 | @forever21 | 948 | 2,842 | 25,918 | 2,082 | 765 | 792 |
| H&M | @hm | 587 | 3,352 | 15,808 | 216 | 3 | 3 |
| Miu Miu | @miumiu | 680 | 2,407 | 21,054 | 0 | 9 | 9 |
| Monsoon | @monsoon | 314 | 3,084 | 16,747 | 82 | 17 | 20 |
| Patagonia | @patagonia | 12 | 383 | 716 | 0 | 2 | 2 |
| UGG | @ugg | 555 | 3,494 | 19,150 | 761 | 15 | 15 |
| Wrangler | @wrangler | 313 | 1,815 | 7,711 | 295 | 6 | 6 |
| | **TOT** | 6,163 | 29,501 | 199,666 | 5,149 | 899 | 953 |

Table 12: Preprocessed corpus files details

No cut-off threshold was set, and the obtained wordlists were saved to one .csv file each, later used as input for comparison against the list of emojis and homoglyphs effectively present in the source data. Emojis and homoglyphs mismatches were only found in the non preprocessed corpora, with the preprocessed ones (both pre-tagged and plain-text versions) faithfully representing both features as they are included in the source data. Consequently the following analysis only takes into account results from the non preprocessed corpora. Table 13 reports details for emojis recognition in the three tools, with column ***Emoji types*** containing the number of types in each wordlist that contain at least one emoji (for ***Source data*** the figure represents the number of unique emojis as counted by the library `emoji`). Consequently e.g. the 991 types identified by AC include ones where a single emoji is present (e.g. 💬; codepoint: U+1F4AC); ones where tokenisation did not split two or more consecutive emojis (e.g. 🌿🤎; codepoints:

U+1FAD2 U+1F90E), and ones where one or more consecutive emojis and a preceding or following string of text were not split (e.g. 'go🪞'; emoji codepoint: U+1FA9E).

| | Emoji types | Tokens | Unrecognised emojis | Invalid emoji types |
|---|---|---|---|---|
| Source data | 401 | 5,149 | — | — |
| AntConc | 991 | 4,606 | 133 | 723 |
| LancsBoxX | 396 | 5,245 | 75 | 70 |
| SketchEngine | 386 | 5,021 | 20 | 5 |

Table 13: 'Emoji types' identified by the corpus tools

Various degrees of infidelity to the source data emerge from these results. Whereas AC does not split emoji characters or sequences from strings of text (e.g. ***loveislove🌈💝***) and splits emoji sequences into their components (see 4.1), both LBX and SkE appear to tokenise the vast majority of emojis present in the source data. This, at least, judging from the column Emoji types. However – as hinted by the remaining columns – a number of emojis present in the source data are not identified by neither LBX nor SkE (column ***Unrecognised emojis***); and a number of emojis tokenised by LBX and SkE do not actually appear in the source data, but rather are the result of splitting valid sequences into different characters or invalid sequences (column ***Invalid emoji types***). For reasons of space, Table 14 only reports details for those emoji types that are recognised by SkE but not present in the source data.

| Emoji type | Codepoints | Source data | AntConc | SketchEngine | LancsBoxX |
|---|---|---|---|---|---|
| ⚧ | U+26A7 | n | y | y | y |
| 🏳️ | U+1F3F3 U+FE0F | n | n | y | n |
| 🏼 | U+1F3FC | n | n | y | y |
| 😵 | U+1F635 | n | y | y | y |
| 🛼🛼 | U+1F6FC U+1F6FC | n | n | y | n |

Table 14: Emoji types identified by SketchEngine but not appearing in the source data

The additional details included in Table 14 report whether the emoji types are also identified by AC and LBX (with a yes y, or a no n). A closer look at the source data and the SkE corpus indicates the causes of these erroneous emoji types: ⚧ and 🏳️ are the result of splitting 🏳️‍⚧️ (codepoint: U+1F3F3 U+FE0F U+200D U+26A7 U+FE0F); 🏼 is the result of splitting 🤝🏼 (codepoint: U+1F91D U+1F3FC); 😵 is the result of splitting 😵‍💫 (codepoint: U+1F635 U+200D U+1F4AB); and 🛼🛼 results from the non separation of the two occurrences of 🛼 (codepoint: U+1F6FC), which in the source data appear followed by a string of text without separating whitespace ('🛼🛼down'), apparently inhibiting proper tokenisation. The presence of one or more emojis next to one or more strings of text – without whitespace separations – is arguably also the cause of 12 similarly structured invalid emoji types tokenised by LBX, with the remaining mismatches being the result of multiple emoji characters or sequences joined together (n=7); emoji components split from their original sequences (n=47); and emoji characters to which LBX adds either the nonprintable character Zero Width Space (codepoint: U+200B; n=1) or the nonprintable character Word Joiner (codepoint: U+2060; n=3). The lack of normalisation in both AC and LBX also results in additional differences in types when homoglyphs are involved (together with the ones discussed in 4.2). For example, whenever in the source data one or more emoji appears next to one or more homoglyph (with no whitespace separation), AC treats the sequence "emoji+homoglyph" as one single token, whereas LBX splits the two elements as two different ones - as exemplified in Fig.3, with the tokens from AC on the left, and the ones from LBX on the right.

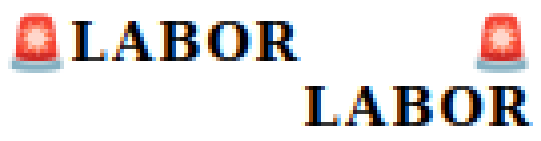

Figure 3: Emoji+homoglyphs tokenisation in AntConc and LancsBoxX

Non-normalisation of homoglyphs also produces further issues for the analysis of the corpus: two examples should clarify the implications. The first concerns the brand name ***Chloé***, present in the source data and in the non preprocessed corpora in both AC and LBX with two different forms: with the decomposed form of the character ***é*** (i.e. ***e*** and the accent as two separated characters; codepoints: U+0065 U+0301) counted as type with frequency = 11; and with the character ***é*** as a single precomposed form (codepoint: U+00E9) counted as type with frequency = 256. The latter version is instead the only one present in the preprocessed files, resulting in the type ***Chloé*** appearing 267 times. The second example represents cases where homoglyphs appear in combination with already normalised characters, such as '**PRESIDENT'S**' (where the standard character "Right Single Quotation Mark" - codepoint: U+2019 – is used in place of the apostrophe – codepoint: U+0027 -, treated by all three tools as apostrophe). In such cases both AC and LBX treat the whole string as a type, rather than splitting it into the types ***president*** and ***'s***. SkE as well, and despite normalisation, creates a separate type for the normalised string ***president's*** rather than adding the occurrences to the two types ***president*** and ***'s***, suggesting that once normalisation is applied the normalised version is not retokenised, leading to the arising of mismatches in types and tokens calculations.

## 5. Conclusions

Despite UTF-8 being the most widely used encoding on the web and the preferred choice in corpus linguistics, its complexity and the increasing number of characters have not kept pace with advances in former; and issues related to characters reading identified 20 years ago (McEnery & Xiao, 2005) persist today in new forms. Emojis and homoglyphs, now prevalent in textual data, represent (one) current incarnation of the source of such mismatches, as seen in the difficulties three popular corpus tools (AntConc, LancsBoxX, and SketchEngine) face in accurately tokenising them. The result is a lack of fidelity of the corpus to the source data, inhibiting researchers to effectively analyse the language they wish to investigate. Preprocessing emojis and homoglyphs in corpus files prior to the compilation of a corpus can ensure that mismatches with the source data are avoided (or limited to a minimum), guaranteeing at the same time that researchers are able to employ the tool they are most confident with – or the one that serves their purpose best. This requires the need to engage with technical aspects but, as daunting as this may seem, knowledge in (corpus) linguistics represents a privileged starting point. Preprocessing is in fact always guided by linguistic knowledge (or rather, it is an integral part of linguistics), and dictated by the linguistic purpose the data must fulfil. Whether the emoji ™ should be transliterated into its standard CLDR name ***trade mark*** or converted into the string ***TM*** (see 3.3) is tied to how the data is going to be linguistically investigated; unifying all possible forms of the accented ***é*** so that all the occurrences of the abstract character as perceived by the researcher are equally "perceived" by the computer (see 4.3) directly speaks to the analysis to be conducted. In doing so, preprocessing does not introduce any additional or unwanted linguistic-related value: CLDR names should in effect be intended as 'ready-made unique emojis labels' – with the added benefit of being intelligible from a plain-text only version of the source data – ensuring that emojis are correctly identified, and that source data can be reconstructed from the preprocessed files. Homoglyphs normalisation can unify characters that are content-wise equivalent or indistinguishable from one another to the human eye, thus reversing the computer-to-human priority involved in the processing of textual data, while bringing researchers closer to "what the writer saw while writing". And, at last, metadata (e.g. in the form of XML tags) guarantees the possibility of perusing the original forms at any time. Overlooking the preprocessing of emojis and homoglyphs may result in consequences that go beyond mere quantitative aspects (e.g. collocates and keywords), affecting qualitative approaches and theoretical underpinnings. The splitting of the "Transgender Flag" 🏳️‍⚧️ (see 4.3) into ⚧ and 🏳 – resulting in the original emoji to be missing in the corpus – directly impacts investigations in i.a. gender studies and (Critical) Discourse Analysis. Similarly the splitting of skin tones and gender variants from the affected emojis precludes researchers from taking into account the role that these play in users' perception (Robertson et al., 2021). The incorrect tokenisation – even when normalisation is automatically applied by the corpus tool – of the string ***president's*** when homoglyphs are used (see 4.3) causes the genitive to be left unrecognised, with consequent repercussions on studies in i.a. syntax, grammar, and lexis. Implementing full support for emojis and homoglyphs in the three tools taken into consideration – and, possibly, any other tool – arguably requires the development and implementation of solutions that are dependant on how a tool was devised in the first place. These solutions would also require subsequent maintenance, increasing the workload of developers. The preprocessing procedures suggested in this paper consequently represent a viable and rapid way to overcome this situation, ensuring compatibility and repeatability across different operating systems and tools, while making full use of text-oriented corpus methods for the analysis of features with high degrees of visual components, without any information loss. How to process non-standard Unicode characters is nonetheless a task that linguists must share with computer sciences knowledge and practitioners, as without the latter we may risk reinventing the wheel, and without the former the task may become detached from the communicative and social dimensions of language.